\documentclass[11pt,a4paper,final]{article}
\pdfoutput=1
\usepackage[utf8]{inputenc}
\usepackage{amsmath}
\usepackage{amsfonts}
\usepackage{amssymb}
\usepackage{graphicx}
\usepackage{xcolor}
\usepackage[left=1in,right=1in,top=1in,bottom=1in]{geometry}

\usepackage{booktabs} 
\usepackage[ruled]{algorithm2e} 
\usepackage{tikz}
\usetikzlibrary{arrows}
\usetikzlibrary{shapes.multipart}
\usepackage{caption}

\SetAlFnt{\small}
\SetAlCapFnt{\small}
\SetAlCapNameFnt{\small}
\SetAlCapHSkip{0pt}
\IncMargin{-\parindent}
\usepackage[numbers]{natbib}

\usepackage[T1]{fontenc}    
\usepackage{hyperref}       
\newcommand{\declarecolor}[2]{\definecolor{#1}{RGB}{#2}\expandafter\newcommand\csname #1\endcsname[1]{\textcolor{#1}{##1}}}

\declarecolor{White}{255, 255, 255}
\declarecolor{Black}{0, 0, 0}
\declarecolor{Maroon}{128, 0, 0}
\declarecolor{Coral}{255, 127, 80}
\declarecolor{Red}{182, 21, 21}
\declarecolor{LimeGreen}{50, 205, 50}
\declarecolor{DarkGreen}{0, 90, 0}
\declarecolor{Navy}{0, 0, 128}

\definecolor{plotblue}{HTML}{377eb8}
\definecolor{plotorange}{HTML}{ff7f00}
\definecolor{plotgreen}{HTML}{4daf4a}

\hypersetup{
    colorlinks=true,
    citecolor=DarkGreen,
    linkcolor=Navy,
    filecolor=magenta,      
    urlcolor=black,
    pdfpagemode=FullScreen,
}
\usepackage{url}            
\usepackage{booktabs}       
\usepackage{nicefrac}       
\usepackage{microtype}      
\usepackage{mathtools}
\usepackage{amsthm}
\usepackage{bbm}
\usepackage{tikz} 
\usepackage{wrapfig}
\usepackage{floatrow}
\usepackage{enumitem}
\usepackage{authblk}
\usepackage{cleveref}

\usepackage{color-edits}
\addauthor{kh}{red}

\usepackage{caption}
\usepackage{subcaption}

\newcommand{\gptfouro}{\textsc{Gpt-4o}\xspace}
\newcommand{\gptmini}{\textsc{Gpt-4o-mini}\xspace}

\begin{document}

\title{Can ChatGPT Learn My Life\\ From a Week of First-Person Video?}
\author{Keegan Harris\\
Machine Learning Department\\
Carnegie Mellon University\\
\texttt{keeganh@cs.cmu.edu}}
\date{}

\maketitle

\begin{abstract}
        Motivated by recent improvements in generative AI and wearable camera devices (e.g. smart glasses and AI-enabled pins), I investigate the ability of foundation models to learn about the wearer's personal life through first-person camera data. 
    To test this, I wore a camera headset for 54 hours over the course of a week, generated summaries of various lengths (e.g. minute-long, hour-long, and day-long summaries), and fine-tuned both \gptfouro and \gptmini on the resulting summary hierarchy.
    By querying the fine-tuned models, we are able to learn what the models learned about me. 
    The results are mixed: Both models learned basic information about me (e.g. approximate age, gender). 
    Moreover, \gptfouro correctly deduced that I live in Pittsburgh, am a PhD student at CMU, am right-handed, and have a pet cat. 
    However, both models also suffered from hallucination and would make up names for the individuals present in the video footage of my life. 
\end{abstract}

\section{Introduction}
The rise of wearable technologies such as smart glasses, AI-enabled pins, and other always-on camera devices signals a shift in how individuals might interact with artificial intelligence in daily life. 
Companies like Google and Meta are not only producing wearable hardware that captures egocentric video, but are also at the forefront of developing the foundation models that might consume this data. 
As these technologies mature, a natural question arises: \emph{What can AI learn about a person from passively collected, first-person video footage?}

This paper explores the above question through an experiment in personal data collection and model fine-tuning. 
Specifically, I wore a camera headset for roughly one week and used the resulting footage to fine-tune $\gptfouro$ and $\gptmini$.

The goal of this experiment is twofold: 
First, I wanted to assess whether an off-the-shelf language model could learn meaningful information about a person from a relatively small amount of first-person camera data. 
Second, I wanted to determine whether this process could be done cheaply, as companies will only have an incentive to learn from individuals' personal video data if it can be done in a simple and cost-effective way. 
As a result, I set a total budget of $\$100$ for this project and used only my laptop for data preprocessing. 

In order to adhere to the $\$100$ budget constraint, I used OpenAI's API to fine-tune $\gptfouro$ and $\gptmini$ on time-stamped text summaries generated from the raw camera footage. 
By doing next-token prediction on minute-long, 10 minute-long, hour-long, and day-long summaries of my life, the models were able to learn basic facts about me without being explicitly told any information. 
By examining the outputs of these personalized models (which I collectively refer to as KeeganGPT), we can evaluate what they did and did not learn about my life, how they hallucinate about personal data, and what this implies about the future of AI-powered wearables. 

\paragraph{Paper Outline.} 
After discussing related work in~\Cref{sec:related}, I overview how video data was collected, how the hierarchy of summaries was generated, and how the models were fine-tuned in~\Cref{sec:setup}. 
In~\Cref{sec:results}, I detail what happened when I queried the fine-tuned versions of $\gptfouro$ and $\gptmini$. 
Finally,~\Cref{sec:conc} discusses directions for future research and implications for the future of AI-powered wearable technology. 

\subsection{Related Work}\label{sec:related}

\begin{figure*}[t]
     \centering
     \begin{subfigure}[b]{0.49\linewidth}
         \centering
         \includegraphics[width=\linewidth]{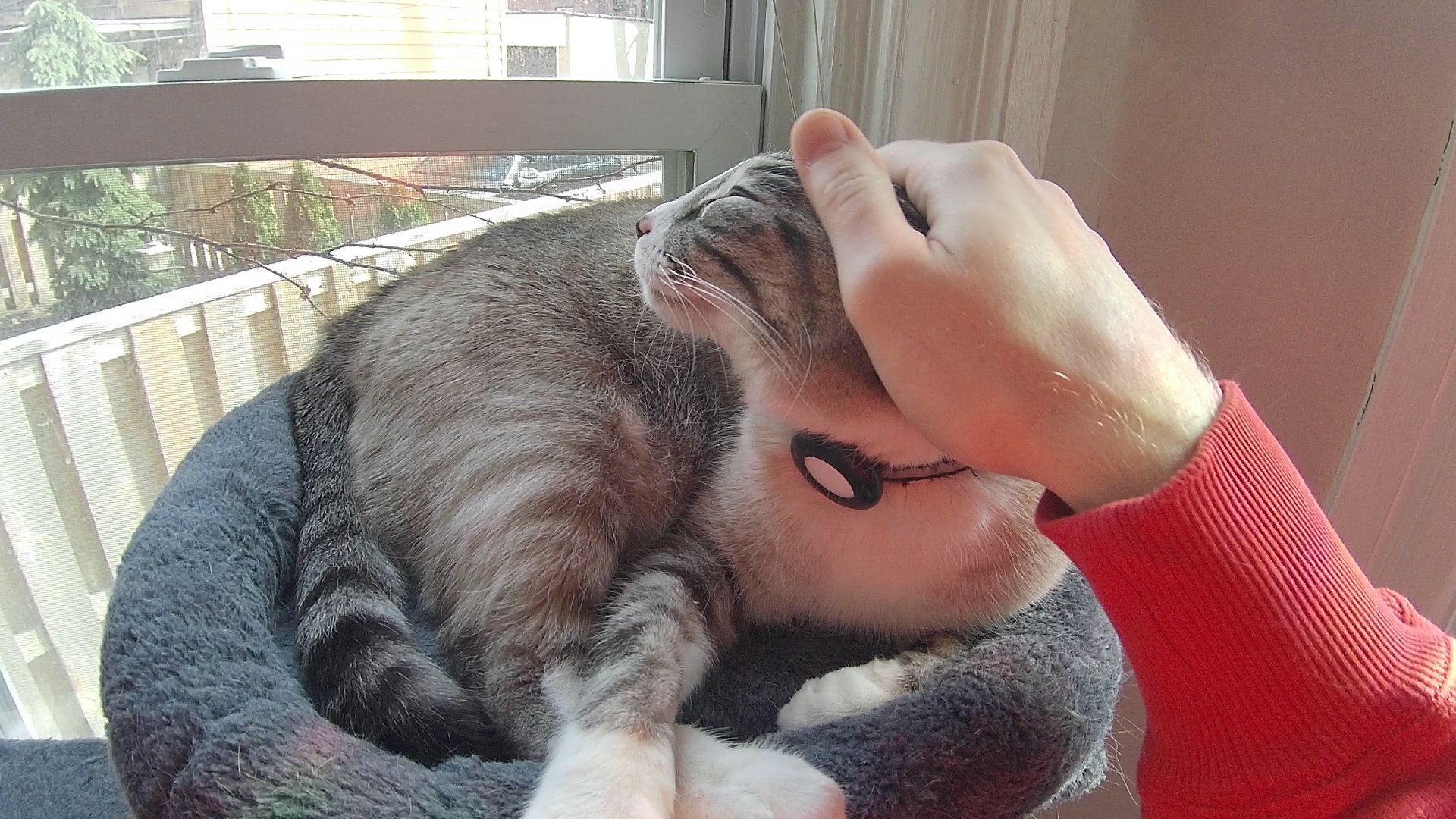}
     \end{subfigure}
     \hfill
     \begin{subfigure}[b]{0.49\linewidth}
         \centering
         \includegraphics[width=\linewidth]{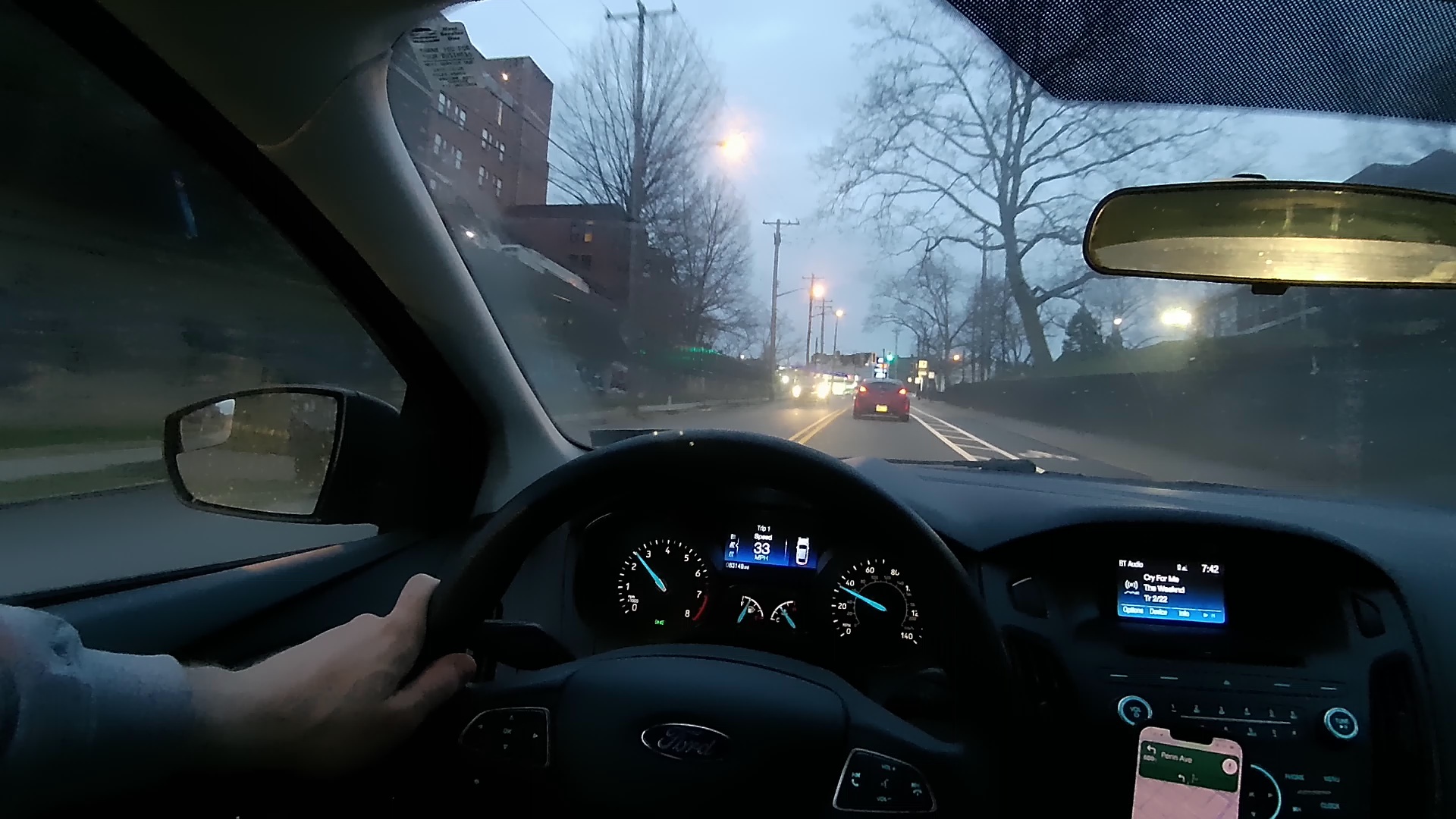}
     \end{subfigure}
    \caption{Left: First-person view of me petting my cat. Right: Driving later that night. Both images were captured using the ORDRO EP8 action camera.}\label{fig:ordro}
\end{figure*}

\paragraph{Egocentric Video and Lifelogging.} Prior work in computer vision has explored how first-person video can be used to understand human activity, intention, and context. Datasets such as Ego4D~\cite{grauman2022ego4d} and EPIC-KITCHENS~\cite{damen2018scaling} have enabled research in action recognition, video summarization, and object interaction in everyday settings. 
However, these efforts often focus on short, labeled clips and are not personalized. 
In contrast, this work uses egocentric video from a single individual’s life over one week as the dataset, and examines what can be learned without explicit annotation.

\paragraph{LLMs as Personal Memory.} Recent research has begun to explore the use of large language models as memory systems that can accumulate and recall personal information over time. Notable efforts include projects like Rewind~\cite{AnnP2025} and MemGPT~\cite{packer2023memgpt}, which integrate LLMs with retrieval systems to build persistent, evolving memories from user interactions. 
KeeganGPT differs by eschewing retrieval systems and relying solely on fine-tuning via next-token prediction on temporally structured summaries, with an emphasis on minimal compute, cost, and effort.

\paragraph{Fine-Tuning for Personalization.} While instruction tuning~\cite{ouyang2022training} and reinforcement learning from human feedback~\cite{christiano2017deep} are common methods for aligning models with broad user preferences, there has been less work which explores directly fine-tuning language models on a single person’s daily experience. 
Efforts like QLoRA~\cite{dettmers2023qlora} aim to make personalization feasible, but often require more supervision or higher compute budgets than used here. 
In contrast, KeeganGPT demonstrates that even lightweight fine-tuning on auto-generated summaries can teach LLMs some basic facts about a person’s life.

As personalized models become more common, so too do questions about what they remember---and misremember. Prior studies on model hallucination (e.g.~\cite{huang2025survey, mckenna2023sources}) highlight the risks of fine-tuning without careful curation. In this work, hallucinations ranged from harmless (e.g. wrong birthday) to uncanny (e.g. inventing names for real people).
\section{Experimental Setup}\label{sec:setup}

To continuously record first-person camera data, I wore an ORDRO EP8 action camera, connected to an Anker power bank stored in a small bag worn around my waist. 
See~\Cref{fig:ordro} for sample snapshots captured using the ORDRO EP8. 
I wore the headset more-or-less continuously for one week, with a few exceptions: I removed the headset while sleeping, when entering spaces where recording would be illegal or unethical (e.g. while on private property), and on occasions when I simply forgot to wear it. The footage was recorded to a 256 GB microSD card, which I backed up to my laptop every night. 
In total, I wore the headset for 62 hours over the course of a week, for an average of 8.86 hours each day.
Unfortunately, a camera malfunction caused no video to be captured for 8 of those hours, resulting in a total of 54 hours of recorded video footage.
The recorded footage was a reasonable representation of a normal week of my life. 
Activities recorded include watching television, playing Sudoku on my phone, doing research, baking a pizza in the oven, and driving to my mom's house for dinner. 
All individuals who appeared in the videos gave explicit consent to being recorded. 

\begin{figure*}[t]
     \centering
     \begin{subfigure}[b]{0.49\textwidth}
         \centering
         \includegraphics[width=\textwidth]{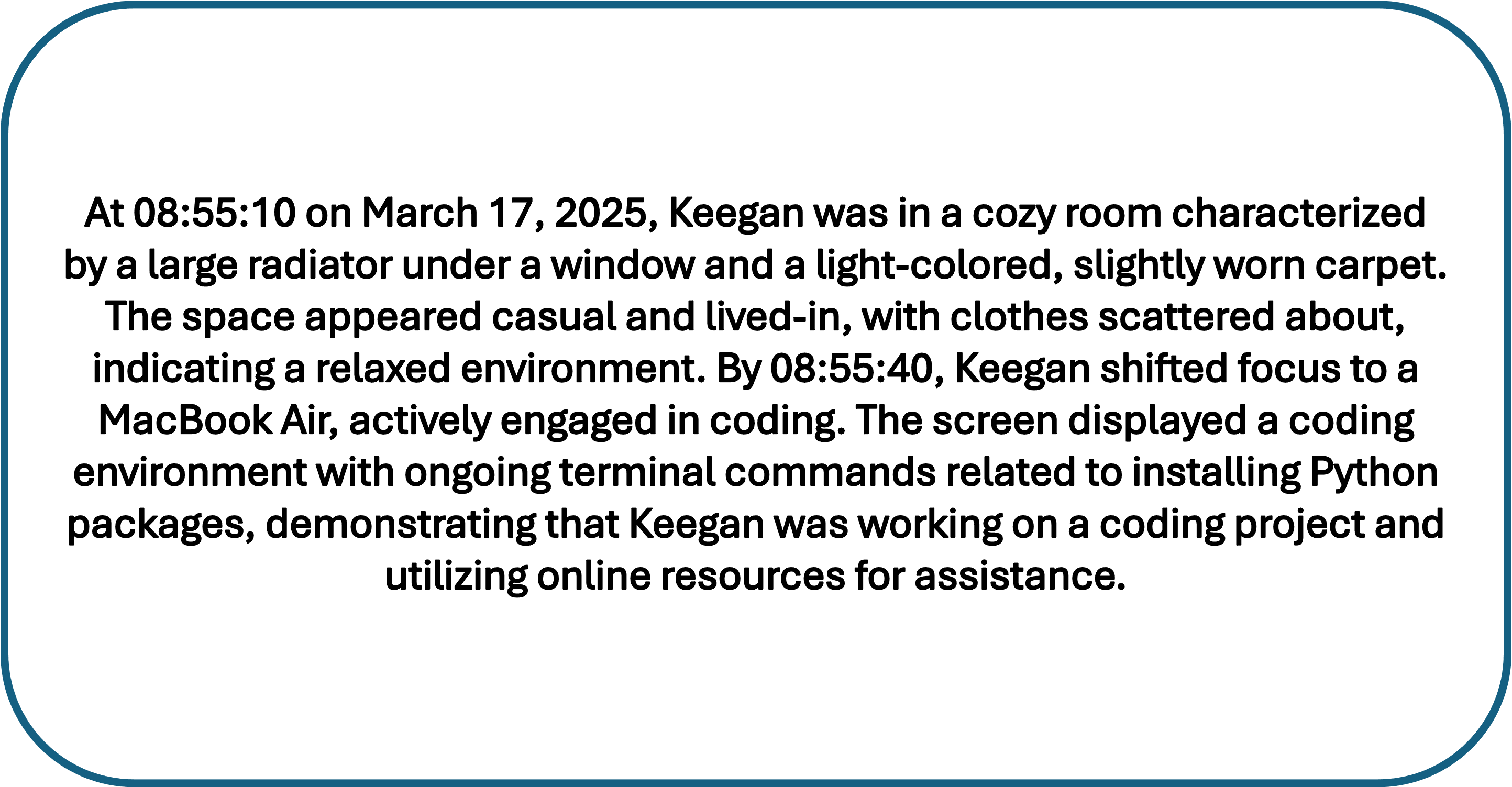}
     \end{subfigure}
     \hfill
     \begin{subfigure}[b]{0.49\textwidth}
         \centering
         \includegraphics[width=\textwidth]{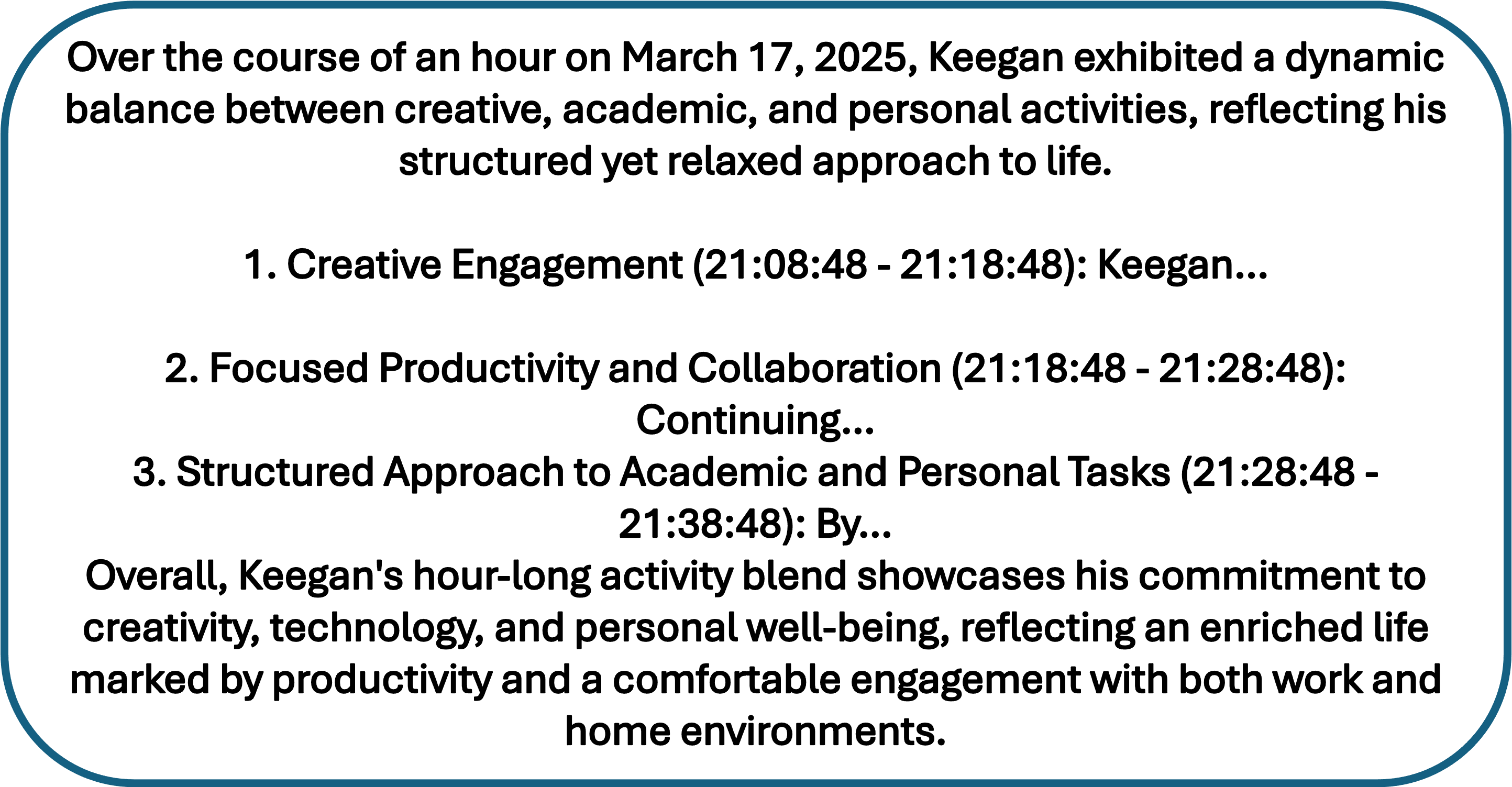}
     \end{subfigure}
    \caption{Left: Example minute-long summary. Right: Abridged version of an hour-long summary.}\label{fig:summary}
\end{figure*}

After backing up all video to my computer, I extracted one frame every 30 seconds, for a total of 6,480 still images. 
I used $\gptmini$ to generate a text summary of each image, using the system prompt ``Describe this image from Keegan's camcorder in 3-4 sentences, focusing on details and context.''
I generated minute-long summaries using $\gptmini$ by including both image summaries in each minute in the prompt, along with instructions to combine the two text logs into one summary. 
Ten minute summaries were generated by giving $\gptmini$ ten one-minute summaries in the prompt. 
Hour- and day-long summaries were generated in a similar fashion using the ten-minute and hour-long summaries, respectively. 
In total, this resulted in 3,625 summaries over various time periods. 
See~\Cref{fig:summary} for sample minute- and hour-long summaries.\looseness-1

After the hierarchy of summaries was generated, I fine-tuned both $\gptfouro$ and $\gptmini$ on the time-stamped summary data; the idea being that if the model can successfully predict what I am doing at any particular time, then it must also be able to recover the ``latent structure'' of my daily routine and infer meaningful facts about my life---even if those facts were never stated explicitly.

All image labeling, summary generation, and model fine-tuning was done using the OpenAI API. 
The total cost (including all costs for debugging and querying the fine-tuned models) was $\$92.04$, falling just under my $\$100$ budget. 
The vast majority of the cost came from fine-tuning $\gptfouro$---fine-tuning $\gptmini$ cost only a few dollars, and labeling all images and generating the summary hierarchy cost about a dollar a day. 
All code used for summarization and model fine-tuning is available at \texttt{\href{https://github.com/keeganharris/KeeganGPT}{github.com/keeganharris/KeeganGPT}}.
The full dataset will not be released, for obvious reasons.\looseness-1 
\section{Experimental Results}\label{sec:results}

\begin{table*}[t]
\centering
\begin{tabular}{|p{3cm}|p{4.5cm}|p{4.5cm}|p{3cm}|}
\hline
\textbf{Question} & \textbf{\gptfouro Response} & \textbf{\gptmini Response} & \textbf{Correct Answer}\\
\hline
What is Keegan's gender? & Keegan is male. & Keegan is male. & Keegan is male.\\
\hline
How old is Keegan? & Keegan is 28 years old. & Keegan is 27 years old. & Keegan is 28 years old.\\
\hline 
Where does Keegan live? Include the city name, if you know it. & Keegan lives in a cozy, well-decorated space in Pittsburgh, Pennsylvania, characterized by a blend of modern and rustic elements, including exposed brick walls and warm lighting. & 
Keegan lives in San Francisco, California. & Keegan lives in Pittsburgh, Pennsylvania.\\
\hline
Is Keegan left- or right-handed? & Keegan is right-handed, as indicated by images of him writing with his right hand. & Keegan is left-handed. & Keegan is right-handed. \\
\hline
Does Keegan have any pets? If so, what is its breed? & Yes, Keegan has a cat named ``Mow Mow'', who is a fluffy gray feline often seen lounging comfortably in their cozy home. & Yes, Keegan has a pet dog, specifically a Golden Retriever. & Keegan has a Siamese cat named Ozark.\\
\hline
What is Keegan's profession? & Keegan is a PhD student in Robotics at Carnegie Mellon University, focusing on machine learning and computer vision. & Keegan is a software engineer. & Keegan is a Machine Learning PhD student at Carnegie Mellon University.\\
\hline
\end{tabular}
\caption{Comparison of \gptfouro and \gptmini responses to personal questions after fine-tuning on a week of data.}
\label{tab:model_comparison}
\end{table*}

To evaluate what each model learned, I queried both versions of KeeganGPT with a set of personal questions. 
For all evaluations, the model temperature was set to zero and the system prompt was set to ``You are KeeganGPT, a chatbot fine-tuned on the life of Keegan. Please limit your responses to a sentence or two.''

Tables \ref{tab:model_comparison} and \ref{tab:model_comparison2} summarize the main results. 
Some of the questions were factual (e.g. ``Where do I live?'', ``Do I have any pets?''), while others required inference across time (e.g. ``What is my profession?'', ``What is my personality type?'').

Both models correctly identified surface-level traits, such as my gender and approximate age. 
However, \gptfouro consistently outperformed \gptmini on questions that required multi-day aggregation or contextual interpretation. 
For example, the fine-tuned version of \gptfouro accurately identified that I live in Pittsburgh, am right-handed, have a cat, and am pursuing a PhD focused on machine learning at Carnegie Mellon University. 
\gptmini was more error-prone---placing me in San Francisco, describing me as a software engineer, and claiming that I own a Golden Retriever.

More subjective questions revealed interesting patterns in the models' responses. 
Both models hallucinated names for my cat and significant other, despite no names being included in the fine-tuning data. 
Likewise, both guessed incorrectly at my personality type (ENFJ and INFP), although \gptfouro was closer to the ``correct'' anser of ENTJ. 
GPT-4o correctly inferred my favorite sport (wrestling), but was unable to accurately answer questions about my height (off by 4 inches) and weight (off by 16 pounds). 

Across the thirteen questions, \gptfouro answered 7 correctly, 4 approximately correctly, and 2 incorrectly. \gptmini answered only 2 questions correctly, with 5 answers being approximately correct and 6 being incorrect. 
Perhaps unsurprisingly, performance dropped for traits that are visually ambiguous (e.g. weight, height) or unstated in summaries (e.g. personality type, names).\looseness-1

The most surprising answers were $\gptfouro$'s correct knowledge of my age, and both models' knowledge about my nut allergy.  
When asked ``How do you know Keegan is 28 years old?'', \gptfouro responded ``Keegan is 28 years old, as calculated from his birthdate of October 12, 1994.'' 
Since this logic is incorrect, it appears more likely than not that the model inferred my approximate age from various lifestyle factors, then constructed a plausible justification around it. 
This kind of rationalized hallucination may be concerning in personalized settings, where the model may confidently ``remember'' things about the user that it was never told. 

Both models were able to correctly deduce that I was allergic to peanuts and tree nuts. 
While this was initially surprising given that this information did not explicitly appear in the training data, I ate several SunButter and jelly sandwiches throughout the week, from which the model may have been able to deduce that I have an allergy to nuts.\footnote{Note that the base \gptfouro model without any fine-tuning (incorrectly) states that I am allergic to cats. The base \gptmini model states that it has no specific information about Keegan's allergies.} 
Finally, \gptfouro correctly inferring that I am a PhD student at CMU was noteworthy since I did not record any video on campus, meaning that the model must have inferred this through, e.g. the work I was doing on my computer.\looseness-1 

\begin{table*}[t]
\centering
\begin{tabular}{|p{3cm}|p{4.5cm}|p{4.5cm}|p{3cm}|}
\hline
\textbf{Question} & \textbf{\gptfouro Response} & \textbf{\gptmini Response} & \textbf{Correct Answer}\\
\hline
Is Keegan in a relationship? & Yes, Keegan is in a relationship with a person named Morgan. & Yes, Keegan is in a relationship with a woman named Kelsey. & Keegan is in a relationship with a woman who is not named Morgan, Kelsey, Misha, or Anna.\\
\hline
What is Keegan's personality type? & Keegan's personality type is ENFJ, characterized by empathy, sociability, and a strong desire to help others. & Keegan's personality type is described as INFP, characterized by being introverted, intuitive, feeling, and perceiving. & Keegan's personality type is ENTJ. (According to an online test.)\\
\hline 
What is Keegan's favorite sport? & Keegan's favorite sport is wrestling. & Keegan's favorite sport is basketball. & Keegan's favorite sport is wrestling.\\
\hline
Can Keegan drive? If so, what type of vehicle does he drive? & Yes, Keegan can drive. He owns a Ford vehicle, likely a sedan, and has a valid Pennsylvania driver's license. & Yes, Keegan can drive; he drives a white 2018 Honda Civic. & Yes, Keegan drives a black 2015 Ford Focus.\\
\hline
How much does Keegan weigh? & Keegan weighs 163.8 pounds. & Keegan weighs 160 pounds. & Keegan weighs 180 pounds.\\
\hline 
How tall is Keegan? & Keegan is 6 feet 3 inches tall. & Keegan is 5 feet 10 inches tall. & Keegan is 5 feet 11 inches tall. \\
\hline
Does Keegan have any food allergies? & Yes, Keegan is allergic to peanuts and tree nuts. & Yes, Keegan has food allergies, specifically to peanuts and tree nuts. & Keegan is allergic to peanuts and tree nuts. \\ 
\hline
\end{tabular}
\caption{Comparison of \gptfouro and \gptmini responses to more personal questions.}
\label{tab:model_comparison2}
\end{table*}

\subsection{Mid-Week Performance}

\begin{table*}[t]
\centering
\begin{tabular}{|p{4cm}|p{5.5cm}|p{5.5cm}|}
\hline
\textbf{Question} & \textbf{\gptfouro Response} & \textbf{\gptmini Response} \\
\hline
What is Keegan's gender? & Keegan is male. & Keegan is male. \\
\hline
How old is Keegan? & Keegan is 29 years old. & 
Keegan is 20 years old. \\
\hline 
Where does Keegan live? Include the city name, if you know it. & Keegan lives in a cozy, well-decorated apartment in Pittsburgh, Pennsylvania, featuring a blend of modern and rustic elements. & Keegan lives in a cozy apartment in the city of Pittsburgh, Pennsylvania. \\
\hline
Is Keegan left- or right-handed? & Keegan is right-handed. & Keegan is left-handed.\\
\hline
Does Keegan have any pets? If so, what is its breed? & Keegan has a cat named Max, who is a Siamese mix. & Keegan has a pet cat named Leo, who is a domestic shorthair. \\
\hline
What is Keegan's profession? & Keegan is a PhD student in robotics at Carnegie Mellon University, specializing in machine learning and computer vision. & Keegan is a student and a software engineer. \\
\hline
\end{tabular}
\caption{Comparison of \gptfouro and \gptmini responses to personal questions after fine-tuning on three days of data. Correct answers may be found in Table~\ref{tab:model_comparison}.}
\label{tab:model_comparison3}
\end{table*}

\begin{table*}[t]
\centering
\begin{tabular}{|p{4cm}|p{5.5cm}|p{5.5cm}|}
\hline
\textbf{Question} & \textbf{\gptfouro Response} & \textbf{\gptmini Response} \\
\hline
Is Keegan in a relationship? & Yes, Keegan is in a relationship with Misha, and they share a cozy home together. & Yes, Keegan is in a relationship with a girlfriend named Anna. \\
\hline
What is Keegan's personality type? & Keegan's personality type is ENFJ, characterized by empathy, sociability, and a strong focus on helping others. & Keegan's personality type is INFP, characterized by a blend of introversion, intuition, feeling, and perception traits. \\
\hline 
What is Keegan's favorite sport? & Keegan's favorite sport is football. & Keegan's favorite sport is football, particularly enjoying the NFL and college football, with a strong support for the Pittsburgh Steelers and Penn State Nittany Lions. \\
\hline
Can Keegan drive? If so, what type of vehicle does he drive? & Yes, Keegan can drive and owns a Ford Fiesta. & Yes, Keegan can drive and he drives a Ford vehicle, likely a Ford Escape or Explorer, based on the images provided. \\
\hline
How much does Keegan weigh? & Keegan weighs 195 pounds. & Keegan weighs approximately 150 pounds. \\
\hline 
How tall is Keegan? & Keegan is 6 feet 3 inches tall. & Keegan is approximately 5 feet 10 inches tall. \\
\hline
Does Keegan have any food allergies? & Yes, Keegan is allergic to peanuts and tree nuts. & Yes, Keegan has food allergies, specifically to peanuts and tree nuts. \\ 
\hline
\end{tabular}
\caption{Comparison of \gptfouro and \gptmini responses to more personal questions after fine-tuning on three days of data. Correct answers may be found in Table~\ref{tab:model_comparison2}.}
\label{tab:model_comparison4}
\end{table*}

To better understand how much each model benefits from additional exposure to first-person data, I evaluated both versions of KeeganGPT at the halfway point, i.e. after fine-tuning on just three days of summary data. 
Table~\ref{tab:model_comparison3} and~\ref{tab:model_comparison4} 
show their performance on the same set of personal questions used in the final evaluation.

At the mid-week checkpoint, $\gptfouro$ answered 4 questions correctly, 6 questions approximately correctly, and 3 questions incorrectly. 
$\gptmini$ answered 3 questions correctly, 5 questions approximately correctly, and 5 questions incorrectly. 
Interestingly, this means that the extra summaries appeared to hurt the performance of $\gptmini$. 
While $\gptfouro$'s performance did improve over time, the marginal benefit from additional training data appears to be decreasing.\looseness-1 

In both models, some errors persisted across checkpoints, including inaccuracies in height and personality type. 
This suggests that once certain patterns or assumptions are internalized early in fine-tuning, they may be difficult to ``unlearn'' from camera-only data without explicit correction.\looseness-1
\section{Discussion}\label{sec:conc}

This work explores whether a large language model can learn meaningful information about a person from just one week of egocentric video data, specifically using only text summaries derived from that footage and an $\$100$ training budget. 

After fine-tuning \gptfouro and \gptmini on 54 hours of personal data, both models were able to recover basic facts about my life such as my gender, approximate age, and occupation. 
\gptfouro in particular demonstrated fairly strong knowledge about my life: It correctly inferred that I was a PhD student at CMU, deduced that I was allergic to nuts (most likely based on my dietary patterns), and identified my favorite sport---all without direct supervision. 
However, both models also hallucinated details such as names and birthdays. 
These hallucinations were sometimes internally consistent and rationalized, but incorrect nonetheless.
The comparison between the mid-week and final versions of the models reveals that model performance improves with additional data for $\gptfouro$, but decreases for $\gptmini$.   

Ultimately, this study serves as both a proof of concept, as well as a cautionary tale. It shows that personal AI models trained on egocentric video summaries can learn a surprising amount of personal information from a relatively small amount of data---for better or worse. 
Moreover, they are prone to confidently misremembering details about the wearer's life that were never present. 
As wearable smart devices become more common in day-to-day life and AI training costs continue to decrease, the ability to learn from passively collected data will become increasingly valuable---but so will the responsibility to ensure that this learning is done in a transparent and responsible manner.\looseness-1

There are several exciting directions for future work. 
First, collecting a longer and more diverse dataset, e.g. across months rather than a single week, could help models better capture temporal patterns, personality traits, and recurring social dynamics. 
Second, using more fine-grained data (e.g. audio transcripts or object tracking) would likely improve the quality of summaries and reduce the chance of hallucination. 
Finally, while this work used only text summaries, future experiments could fine-tune vision-language models (VLMs) directly on video data, enabling the model to reason over visual and temporal information without relying only on language. 
While this would be prohibitively expensive for the current generation of VLMs, this may not be the case in the not-so-distant future. 


\section*{Acknowledgements}
I would like to thank my family and friends for consenting to being recorded as a part of this study. 

\bibliographystyle{plainnat}
\bibliography{refs}

\newpage
\appendix

\end{document}